\documentclass[draft]{agujournal2019}

\usepackage{url} 
\usepackage{soul}
\usepackage{float}
\usepackage{lineno}
\usepackage{natbib}
\usepackage{amsmath}
\usepackage{subcaption}

\usepackage[inline]{trackchanges}

\nolinenumbers
\draftfalse

\journalname{Geophysical Research Letters}

\begin{document}

%
%


\title{Diffusion-Based Rollouts as a Stabilization Mechanism for Long-Horizon Environmental Forecasting}

%
%




\authors{Marina Vicens-Miquel\affil{1,2,3}, Amy McGovern\affil{1,2,3}, Aaron J. Hill\affil{1,3}, Efi Foufoula-Georgiou\affil{4}, and Samuel S. P. Shen\affil{5}}

\affiliation{1}{University of Oklahoma, School of Meteorology, Norman, Oklahoma}
\affiliation{2}{University of Oklahoma, School of Computer Science, Norman, Oklahoma}
\affiliation{3}{NSF AI Institute for Research on Trustworthy AI in Weather, Climate, and Coastal Oceanography, Norman, Oklahoma}
\affiliation{4}{University of California Irvine, Irvine, California} 
\affiliation{5}{San Diego State University, San Diego, California}




\correspondingauthor{Marina Vicens-Miquel}{marinavicensmiquel@gmail.com}



\begin{keypoints}
\item Diffusion based rollouts can act as a stabilization mechanism across tabular and raster environmental forecasts
\item Stabilization does not guarantee forecast fidelity when recursive forecasts lack sufficient predictive conditioning
\item Diffusion benefits vary with recursive instability and the predictive information available during rollout
\end{keypoints}

%
%

%
%


\begin{abstract} 
Extending forecast lead times while maintaining predictive skill remains a major challenge in environmental forecasting. We investigate diffusion-based rollouts as a stabilization mechanism for recursive forecasting using low-dimensional water-level time series and high-dimensional precipitation fields. Across both modalities, diffusion suppresses recursive error growth, with the largest stabilization occurring where deterministic rollouts are most unstable. However, stabilization does not guarantee forecast fidelity. In the water-level experiments, forecasts progressively lose event-level fidelity as the rollout loses access to external predictive information, and trajectory-level comparisons show that diffusion can remain numerically stable while contracting toward central values and exhibiting reduced variability. In the precipitation experiments, which retain conditioning from numerical weather prediction throughout the rollout, diffusion better preserves spatial organization and event-detection skill. Together, these contrasting experiments indicate that diffusion can control recursive error amplification, while its practical benefit also depends on the predictive information available to constrain future evolution.
\end{abstract}

\section*{Plain Language Summary} 
Long-range machine-learning forecasts often become unstable because predictions are repeatedly fed back into the model, allowing small errors to grow over time. We examine whether diffusion-based forecasting can reduce this problem for coastal water levels and spatial precipitation fields. Diffusion consistently reduces recursive error growth, especially when deterministic forecasts become unstable. However, a stable forecast is not necessarily an informative forecast. In the water-level experiments, the models eventually rely only on their own previous predictions and no longer receive information about future meteorological forcing. Under these conditions, diffusion can keep forecasts numerically stable, while the forecast trajectories show increasing contraction toward typical values and loss of important variability. In the precipitation experiments, forecasts continue to receive information from a numerical weather prediction model, and diffusion better preserves realistic spatial organization and event-detection skill. Together, the two applications show that diffusion can stabilize recursive forecasting, but stabilization cannot replace missing predictive information. The results therefore indicate that its practical value depends on both forecast instability and the information available to constrain future evolution.

%
%

%


%
%
%
%

\section{Introduction}

Extending forecast lead times while maintaining high predictive skill is a long-standing challenge in environmental sciences. In many applications, directly predicting states far into the future is substantially more difficult and less skillful than producing short-term forecasts. As a result, many forecasting systems rely on recursive prediction, where models are trained for short lead times and their predictions are repeatedly fed back as inputs to generate subsequent forecasts \citep{pathak2018model, weyn2020improving, scher2019weather}. This practice, commonly referred to as rollouts, enables longer-horizon prediction while leveraging the higher skill typically achieved at shorter lead times, but it leads to the accumulation of error and uncertainty at each iteration. As small prediction errors compound over time, forecast skill often degrades rapidly at longer horizons. These challenges are particularly important in environmental forecasting, where nonlinear and chaotic dynamics can amplify small perturbations across spatial and temporal scales \citep{lorenz2017deterministic}. Despite these limitations, autoregressive rollouts have become standard practice across deep learning-based time series prediction, numerical weather post-processing, and geophysical modeling \citep{pathak2018model, scher2019weather, weyn2020improving, chase2025score, schreck2024community, ni2025huracan}.

The implications of recursive prediction for machine learning models were first formally studied in sequence modeling. Early work identified the discrepancy between training, where models are conditioned on ground-truth inputs, and inference, where they must rely on their own predictions. \cite{bengio2015scheduled} demonstrated how this mismatch can lead to compounding errors during inference, while \cite{ranzato2015sequence} proposed training strategies designed to mitigate long-horizon degradation. Subsequent approaches, including scheduled sampling, curriculum learning, and architectures designed to improve recursive robustness, have sought to reduce rollout instability \citep{talvitie2014model, bengio2015scheduled, ranzato2015sequence}. However, long-horizon recursive forecasting remains highly susceptible to error accumulation and instability, particularly in dynamically complex environmental systems.

Most existing approaches for improving rollout stability focus on architectural modifications or training procedures. Comparatively less attention has been given to explicitly modeling uncertainty during the recursive forecasting process itself. Probabilistic forecasting methods and ensemble approaches explicitly represent predictive uncertainty \citep{gneiting2014probabilistic, lakshminarayanan2017simple, salinas2020deepar, lim2021temporal}, but uncertainty is often treated as an output of the forecasting system rather than as an intrinsic component of recursive prediction. Recently, diffusion models have emerged as a powerful probabilistic framework for high-dimensional prediction problems \citep{ho2020denoising, song2021scorebased, karras2022elucidating}. By iteratively denoising perturbed states, diffusion models provide a mechanism for sampling from learned conditional distributions rather than producing a single deterministic prediction. Although diffusion-based methods have shown promising performance for forecasting and conditional generation \citep{rasul2021autoregressive}, their behavior during repeated recursive application and their potential role in controlling long-horizon error growth remain less well understood.

In this work, we investigate diffusion-based rollouts as a stabilization mechanism for long-horizon environmental forecasting across water-level time series and spatial precipitation fields. We directly compare diffusion and deterministic autoregressive rollouts using matched predictive architectures, datasets, and normalization strategies within each forecasting modality, allowing the effects of diffusion-based recursive prediction to be separated from architectural and data-driven differences. Importantly, we distinguish rollout stability from forecast fidelity. A forecasting system may remain numerically stable while progressively losing variability, extremes, or other predictive structure, and reductions in aggregate error therefore do not necessarily imply that long-horizon forecasts remain informative.

Our results show that diffusion consistently moderates recursive error growth, while the practical consequences of this stabilization vary across forecasting regimes. Across the three water-level stations, relatively stable rollouts show modest diffusion benefits, whereas stronger recursive degradation is accompanied by greater loss of event-level fidelity. Diffusion can keep these forecasts numerically bounded, but cannot recover future forcing that is absent from the inputs. In contrast, the precipitation experiments retain external numerical weather prediction guidance throughout recursion and show stronger preservation of spatial organization and event-detection skill. These contrasting settings indicate that both recursive instability and available predictive information are important when interpreting diffusion-based stabilization.

By reframing diffusion models as stabilization tools rather than uniform accuracy enhancers, this work provides a framework for interpreting both the benefits and limits of diffusion-based recursive forecasting. The contributions of this work are threefold. First, we systematically evaluate diffusion-based rollouts as an explicit stabilization mechanism across multiple environmental forecasting regimes. Second, we show that diffusion-induced stabilization is regime dependent and varies with both deterministic recursive degradation and the predictive information available during rollout. Third, we demonstrate that rollout stability and forecast fidelity must be evaluated separately, because numerical stabilization can occur either with retained predictive structure or alongside progressive loss of variability and event-level information.

\section{Datasets}
To evaluate diffusion-based rollouts across contrasting environmental forecasting settings, experiments are conducted on low-dimensional water-level time series and high-dimensional raster precipitation datasets with different degrees of external predictive conditioning.

\subsection{Low-Dimensional Tabular Time Series}
Low-dimensional experiments use publicly available cleaned and gap-filled National Oceanic and Atmospheric Administration (NOAA) tide gauge water-level datasets introduced by \cite{vicens2024exploring}, sampled at 6-minute temporal resolution. The dataset includes three Texas Gulf Coast stations, Bob Hall Pier, Rockport, and Port Isabel. The stations exhibit distinct water-level variability and predictability associated with differences in coastal geometry and environmental forcing, providing contrasting regimes for evaluating recursive forecast behavior and stability.

Each station dataset spans five years selected for data completeness \citep{vicens2024exploring}. The three oldest years are used for training, the fourth for validation, and the final year for testing (\ref{appendix:years}). Deterministic models use the validation year for early stopping, whereas diffusion models are trained without validation-based early stopping, consistent with the adopted diffusion training framework \citep{chase2025score, karras2022elucidating}.

All variables are independently standardized using z-score normalization:
\begin{equation}
x_{\mathrm{norm}} = \frac{x-\bar{x}}{s},
\end{equation}

where $\bar{x}$ and $s$ denote the sample mean and sample standard deviation, respectively, computed from the training data and applied consistently to all dataset splits.

\subsubsection{Target and Inputs}

Consistent with \cite{vicens2024exploring}, both the deterministic and diffusion models predict surge rather than total water level, where surge is defined as the observed water level minus the harmonic tidal prediction. The target is the surge 12 hours ahead, with total water level reconstructed by adding the tidal component back to the predicted surge. To predict this target, model inputs consist of the current surge value and lagged surge observations at 1-hour intervals up to 6 hours prior. Although the dataset introduced by \cite{vicens2024exploring} also includes alongshore and cross-shore wind components under a perfect-prog configuration, these variables are excluded here to avoid introducing unrealistically perfect future forcing during recursive rollouts. Consequently, the tabular experiments recursively extend the 12-hour prediction using water-level history alone, without access to future meteorological information.

\subsection{High-Dimensional Raster Data}\label{section_raster_data}
High-dimensional experiments use precipitation datasets constructed following \cite{vicens2025diffusion}. Three geographically distinct regions within the continental United States (CONUS), centered over Oklahoma, Washington, D.C., and Oregon, are evaluated using $512 \times 512$ grids at 1 km spatial resolution. These regions were selected to represent diverse precipitation regimes and geographic influences across CONUS. The dataset spans March 2021 through February 2025, with the first two years used for training, the third year for validation, and the final year for testing. As in the tabular time series experiments, diffusion models are trained without validation-based early stopping. For evaluation, we follow the sampling strategy of \cite{vicens2025diffusion}, using two forecast initializations per day rather than full hourly coverage because of the computational cost of diffusion inference on high-resolution raster data. Initialization hours rotate monthly to sample the diurnal cycle across the evaluation period.

\subsubsection{Target and Inputs}
The model produces a 1-hour-ahead precipitation forecast by predicting the residual between the observed precipitation and the corresponding HRRR forecast:
\begin{equation}
\text{Target} = \text{MRMS}(t+1) - \text{HRRR}_{f01}(t),
\label{eq:residual}
\end{equation}
where $\text{MRMS}(t+1)$ is the observed precipitation one hour ahead and $\text{HRRR}_{f01}(t)$ is the corresponding one-hour HRRR forecast.

The input configuration follows the hybrid framework of \cite{vicens2025diffusion}. Predictors include the current Multi-Radar Multi-Sensor (MRMS) Quantitative Precipitation Estimate (QPE) \citep{Zhang2016}, the two preceding hourly MRMS observations, and High-Resolution Rapid Refresh (HRRR) \citep{Dowell2022} total precipitation forecasts at one- and two-hour lead times (f01 and f02). HRRR forecasts are interpolated from 3 km to the 1 km MRMS grid. Additional predictors include latitude, longitude, and cyclical temporal encodings for hour, day of year, and month. Unlike the tabular configuration, the raster framework therefore includes externally generated forecast information that can be advanced through the recursive rollout. All raster inputs are standardized using the z-score normalization defined in Equation~\ref{eq:znorm}, with statistics computed from the training data and applied consistently to all dataset splits.

\section{Methodology}
We compare deterministic and diffusion-based recursive forecasting across the tabular and raster experiments. Within each modality, both approaches use the same predictive backbone, input data, and recursive forecasting structure. The diffusion approach differs from its deterministic counterpart through noise-conditioned training and iterative denoising during prediction.

\subsection{Forecasting Setup and Rollouts}

Multi-step forecasts are generated recursively in both tabular and raster settings by incorporating model predictions into the input state for subsequent forecasts. Repeated application of this procedure extends the forecast horizon and allows the evolution of recursive error to be evaluated. For tabular forecasting, each predicted surge value is incorporated into the dynamic input state and the lagged values are shifted accordingly before generating the next prediction. As the rollout progresses, observed surge values are progressively replaced by model-generated values until the dynamic input state consists entirely of previous predictions. The rollout therefore becomes fully autoregressive without external future forcing.

For raster forecasting, each predicted precipitation field is incorporated into the sequence of precipitation inputs for the subsequent forecast. The HRRR conditioning fields are simultaneously advanced in forecast lead time, such that f01--f02 becomes f02--f03, then f03--f04, and so forth. Spatial and temporal features are updated consistently with the valid forecast time. Unlike the tabular setting, the raster forecasts therefore remain conditioned on external forecast information throughout the rollout. Purely data-driven precipitation rollouts without HRRR conditioning were previously shown to become rapidly unstable \citep{vicens2025diffusion}; consequently, the present raster experiments focus on diffusion-based stabilization when external predictive information remains available.

\subsection{Model Configuration}

The diffusion models use the Elucidated Diffusion Model (EDM) framework \citep{karras2022elucidating}, which formulates diffusion through a continuous range of noise levels and a noise-conditioned denoising model. During training, targets are perturbed with varying levels of noise and the model learns to recover the corresponding clean states. During inference, predictions are generated by progressively denoising an initially noisy state through a sequence of decreasing noise levels.

\subsubsection{Low-Dimensional Configuration}

Low-dimensional experiments use the multilayer perceptron (MLP) architecture introduced by \cite{vicens2024exploring}. The deterministic model is trained using mean squared error (MSE):

\begin{equation}
\mathcal{L}_{\text{MSE}}
=
\frac{1}{N}
\sum_{i=1}^{N}
(\hat{x}_i-x_i)^2,
\label{eq_mse}
\end{equation}

where $\hat{x}_i$ and $x_i$ denote the predicted and target surge values, respectively.

The tabular diffusion model uses the same MLP backbone within the EDM framework under an $x_0$ prediction formulation. To account for the range of noise levels encountered during diffusion training, we employ a hybrid loss inspired by \cite{vicens2025diffusion}. The noise-scaled component is defined as

\begin{equation}
\mathcal{L}_{\text{scaled}}
=
\frac{1}{N}
\sum_{i=1}^{N}
\frac{(\hat{x}_i-x_i)^2}
{\sigma_i^2+\epsilon},
\label{eq_tab_scaled}
\end{equation}

where $\sigma_i$ is the sampled diffusion noise level and $\epsilon$ is a small constant for numerical stability. This term places greater emphasis on accurate reconstruction at lower noise levels while reducing the contribution of highly perturbed samples.

The final tabular diffusion loss combines the noise-scaled term with the conventional MSE objective:

\begin{equation}
\mathcal{L}_{\text{tab}}
=
\alpha \mathcal{L}_{\text{scaled}}
+
(1-\alpha)\mathcal{L}_{\text{MSE}},
\label{eq_tab_hybrid}
\end{equation}

where $\alpha=0.8$ controls the relative contribution of the two loss components. This value follows \cite{vicens2025diffusion}, where it was selected through hyperparameter tuning to provide a balance between predictive uncertainty and magnitude accuracy. Location-specific diffusion hyperparameters are summarized in \ref{appendix:diffusion_hparams}.

\subsubsection{High-Dimensional Configuration}

Raster experiments use the dilated attention U-Net architecture introduced by \cite{vicens2025diffusion} and summarized in \ref{appendix:architecture}. The deterministic model directly predicts the precipitation residual, whereas the diffusion model uses the same U-Net backbone within the EDM framework to directly predict the clean precipitation residual, with EDM preconditioning. During inference, a single stochastic realization is generated through iterative EDM sampling and used directly as the forecast, without ensemble averaging.

The deterministic model is trained using the percentile-based weighted mean absolute error (MAE) loss introduced in \cite{vicens2025diffusion}. Let $y$ denote the standardized precipitation residual and $\hat{y}$ the corresponding prediction. Percentile thresholds $p_{10}$, $p_{50}$, $p_{75}$, $p_{90}$, and $p_{97}$ are computed from the empirical distribution of $|y|$ in the training set. These region-specific thresholds define transition points for an intensity-dependent weighting function:

\begin{equation}
\begin{aligned}
w(|y|) =\;&
3.5\, S\!\left((|y|-p_{50})\,150\right)
+5.0\, S\!\left((|y|-p_{75})\,50\right) \\
&+6.0\, S\!\left((|y|-p_{90})\,20\right)
+7.0\, S\!\left((|y|-p_{97})\,10\right) \\
&+0.6\left[1-S\!\left((|y|-p_{10})\,150\right)\right],
\end{aligned}
\label{eq_weight}
\end{equation}

where $S(\cdot)$ denotes the sigmoid function, with the sigmoid scaling parameters selected through hyperparameter tuning. This weighting increases the contribution of larger-magnitude residuals while reducing the influence of very small residual values.

\begin{equation}
\mathcal{L}_{\text{det}}
=
\frac{1}{N}
\sum_{i=1}^{N}
w(|y_i|)\,
|y_i-\hat{y}_i|,
\label{eq_raster_det}
\end{equation}

where $N$ is the number of pixels.

The diffusion raster model uses a hybrid objective combining a noise-scaled MAE term with the same percentile-based intensity weighting:

\begin{equation}
\mathcal{L}_{\text{diff}}
=
\alpha
\frac{1}{N}
\sum_{i=1}^{N}
\frac{|y_i-\hat{y}_i|}
{\sigma_i+\epsilon}
+
(1-\alpha)
\frac{1}{N}
\sum_{i=1}^{N}
w(|y_i|)
|y_i-\hat{y}_i|,
\label{eq_raster_diff}
\end{equation}

where $\sigma_i$ is the sampled diffusion noise level for each training instance, $\epsilon$ is a small constant for numerical stability, and $\alpha=0.8$. The hybrid objective combines noise-dependent reconstruction with intensity-dependent weighting of precipitation residuals.

\subsection{Evaluation Metrics}
Forecast performance is evaluated across rollout lead times using metrics appropriate to each forecasting modality. For the tabular experiments, MAE quantifies errors in predicted water level. For the raster experiments, MAE quantifies precipitation magnitude error and is complemented by the Fractions Skill Score (FSS) and Critical Success Index (CSI), using a rain/no-rain threshold, to assess spatial agreement and event detection.

MAE is defined as

\begin{equation}
\mathrm{MAE}
=
\frac{1}{N}
\sum_{i=1}^{N}
|\hat{y}_i-y_i|,
\label{eq:mae}
\end{equation}

where $\hat{y}_i$ and $y_i$ denote the predicted and observed values, respectively, and $N$ is the number of evaluated samples or grid cells.

FSS measures neighborhood-based spatial agreement and is defined as

\begin{equation}
\mathrm{FSS}
=
1-
\frac{\sum_{i=1}^{N}(F_i-O_i)^2}
{\sum_{i=1}^{N}F_i^2+\sum_{i=1}^{N}O_i^2},
\label{eq:fss}
\end{equation}

where $F_i$ and $O_i$ are the forecast and observed fractions of rainy grid cells, respectively, within a $27\times27$ grid-cell neighborhood centered at grid cell $i$. 

CSI measures event-detection skill at the individual grid-cell level and is defined as

\begin{equation}
\mathrm{CSI}
=
\frac{H}{H+M+FA},
\label{eq:csi}
\end{equation}

where $H$, $M$, and $FA$ denote hits, misses, and false alarms, respectively.


\section{Results and Discussion}

We evaluate diffusion-based rollouts relative to deterministic autoregressive baselines across the tabular and raster forecasting tasks. In addition to aggregate error growth, we examine whether numerical stabilization preserves meaningful forecast variability and structure, and how this behavior depends on the predictive information available during recursion.

\subsection{Tabular Rollout Evaluation}

\begin{figure}[!htbp]
\begin{center}
 \noindent\includegraphics[width=\textwidth]{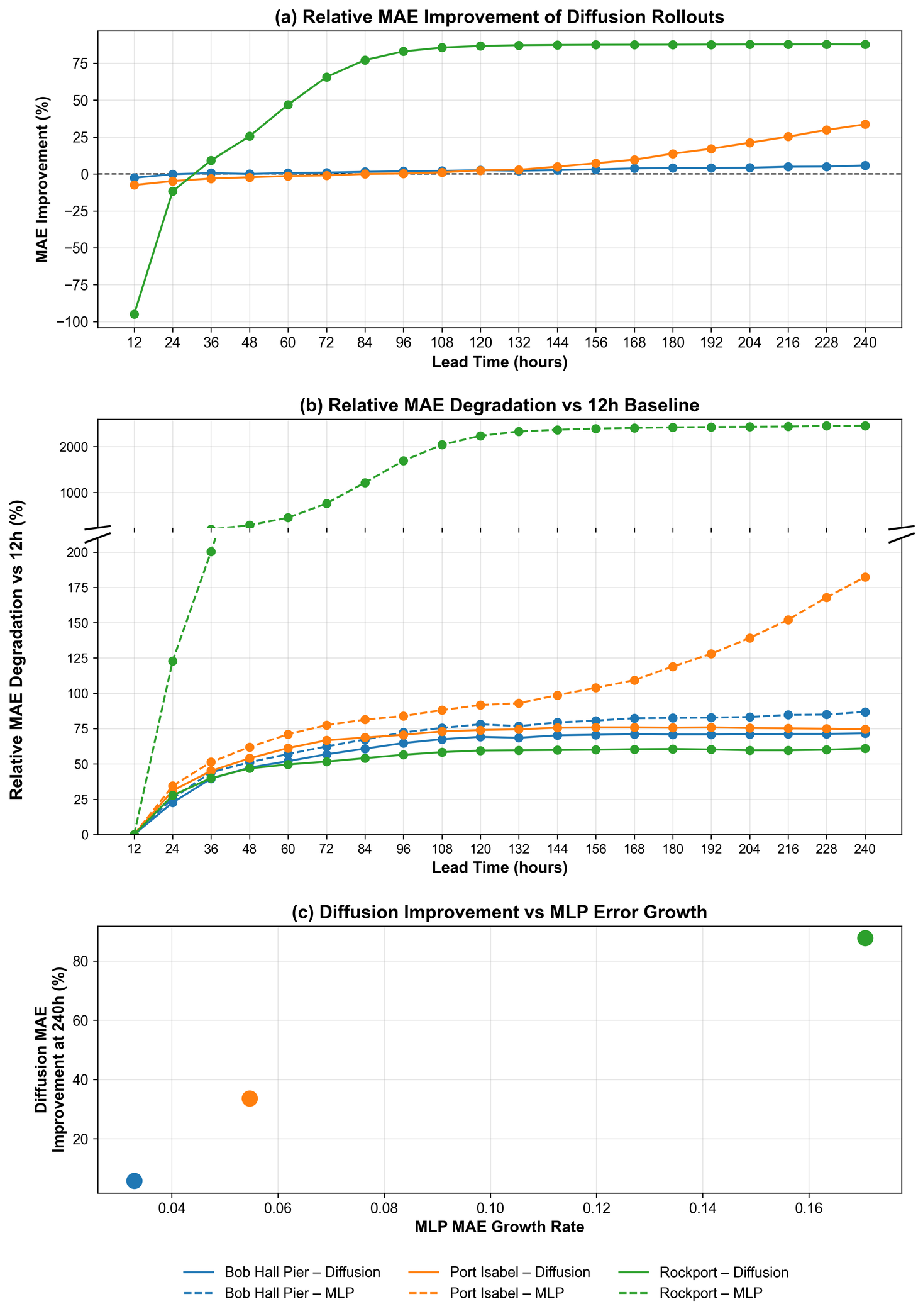}\\
 \caption{Tabular rollout performance across three tide gauge locations. (a) Relative MAE improvement of diffusion rollouts compared with deterministic MLP rollouts across lead times. Positive values indicate improved diffusion performance. (b) Relative MAE degradation with respect to the 12-hour forecast for diffusion (solid) and deterministic MLP (dashed) rollouts. (c) Relationship between deterministic rollout instability and diffusion improvement at 240-hour lead time. Larger deterministic error growth corresponds to greater relative diffusion stabilization.}
 \label{fig_tab_results}
\end{center}
\end{figure}

Figure~\ref{fig_tab_results}a shows the relative MAE improvement of diffusion rollouts compared with the deterministic MLP. At the shortest lead times, diffusion provides little advantage and can slightly increase MAE. As lead time increases, however, diffusion increasingly reduces error relative to the deterministic rollout. The magnitude of this relative improvement differs substantially among stations, with Rockport showing the largest long-horizon gain, Port Isabel an intermediate response, and Bob Hall Pier the smallest. This behavior is further evident in Figure~\ref{fig_tab_results}b. Relative to the 12-hour forecast, diffusion exhibits slower MAE degradation than the deterministic MLP across all three stations. 

\begin{figure}[!htbp]
\begin{center}
 \noindent\includegraphics[width=\textwidth]{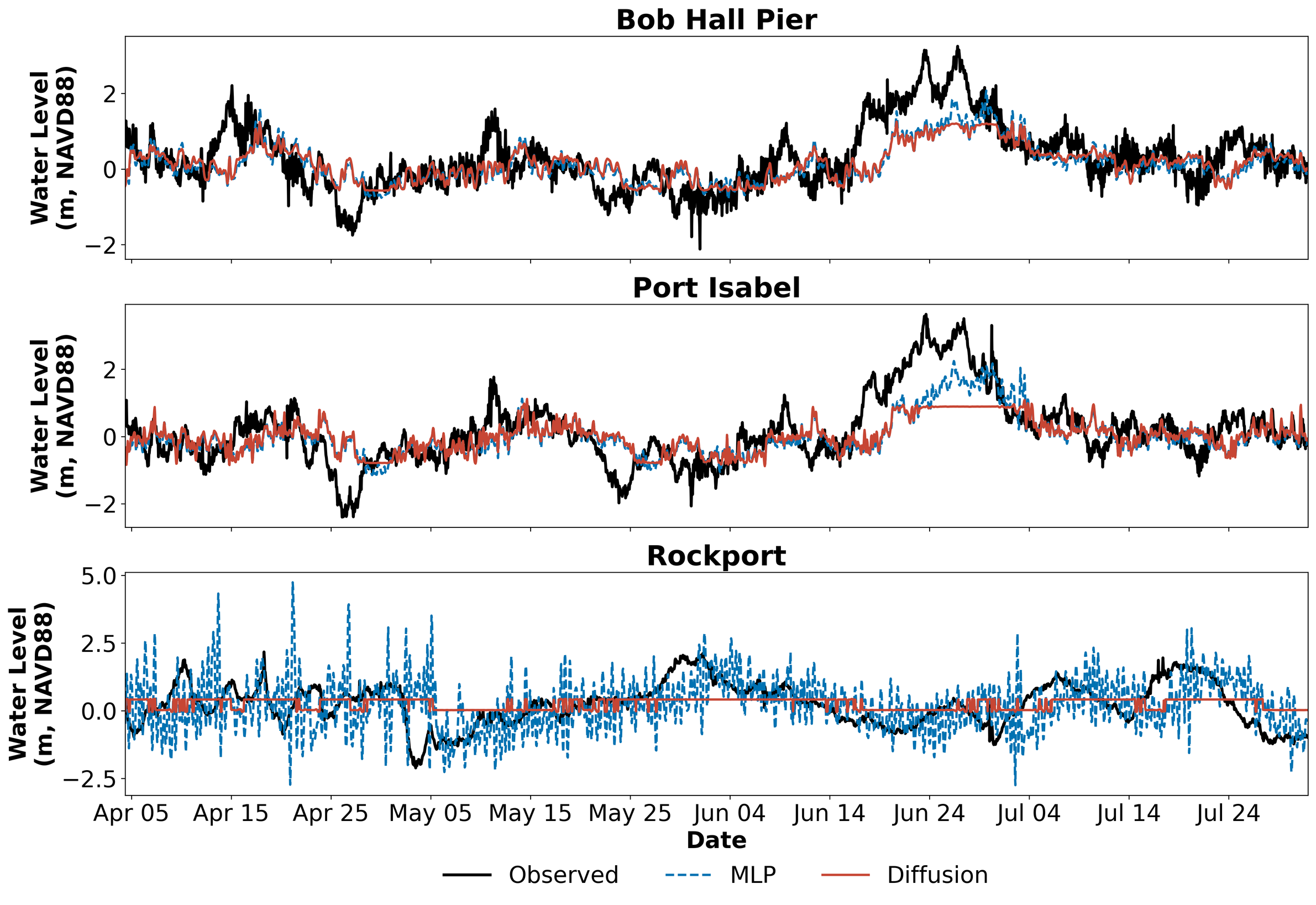}\\
\caption{Observed and predicted water levels at 48-hour lead time for Bob Hall Pier, Port Isabel, and Rockport. The stations exhibit distinct rollout regimes, ranging from comparatively stable behavior at Bob Hall Pier to stronger loss of event-level fidelity at Port Isabel and Rockport. Diffusion remains more numerically constrained, but stabilization does not necessarily imply preservation of forecast variability or event evolution.}
 \label{fig:all_stations_48h}
\end{center}
\end{figure}

Figure~\ref{fig_tab_results}c further shows that larger deterministic error growth corresponds to larger relative diffusion improvements. Importantly, this relationship characterizes stabilization rather than retained forecast skill. A large relative improvement can arise because diffusion remains bounded while the deterministic rollout becomes increasingly unstable, even when neither forecast accurately represents the observed evolution at long lead times. The 48-hour time-series forecasts in Figure~\ref{fig:all_stations_48h} illustrate this distinction and reveal substantial differences across stations. Bob Hall Pier retains the greatest correspondence with the observed evolution, although both models underestimate some variability. Port Isabel exhibits a stronger loss of event amplitude, particularly for higher water-level excursions. At Rockport, both approaches show substantial loss of realistic event evolution, with diffusion remaining more numerically bounded while the deterministic rollout exhibits stronger instability. These results demonstrate that reduced error growth during recursive forecasting does not necessarily imply preservation of realistic forecast evolution.

\begin{figure}[!htbp]
\centering

\begin{subfigure}{\textwidth}
    \centering
    \includegraphics[width=0.72\textwidth]{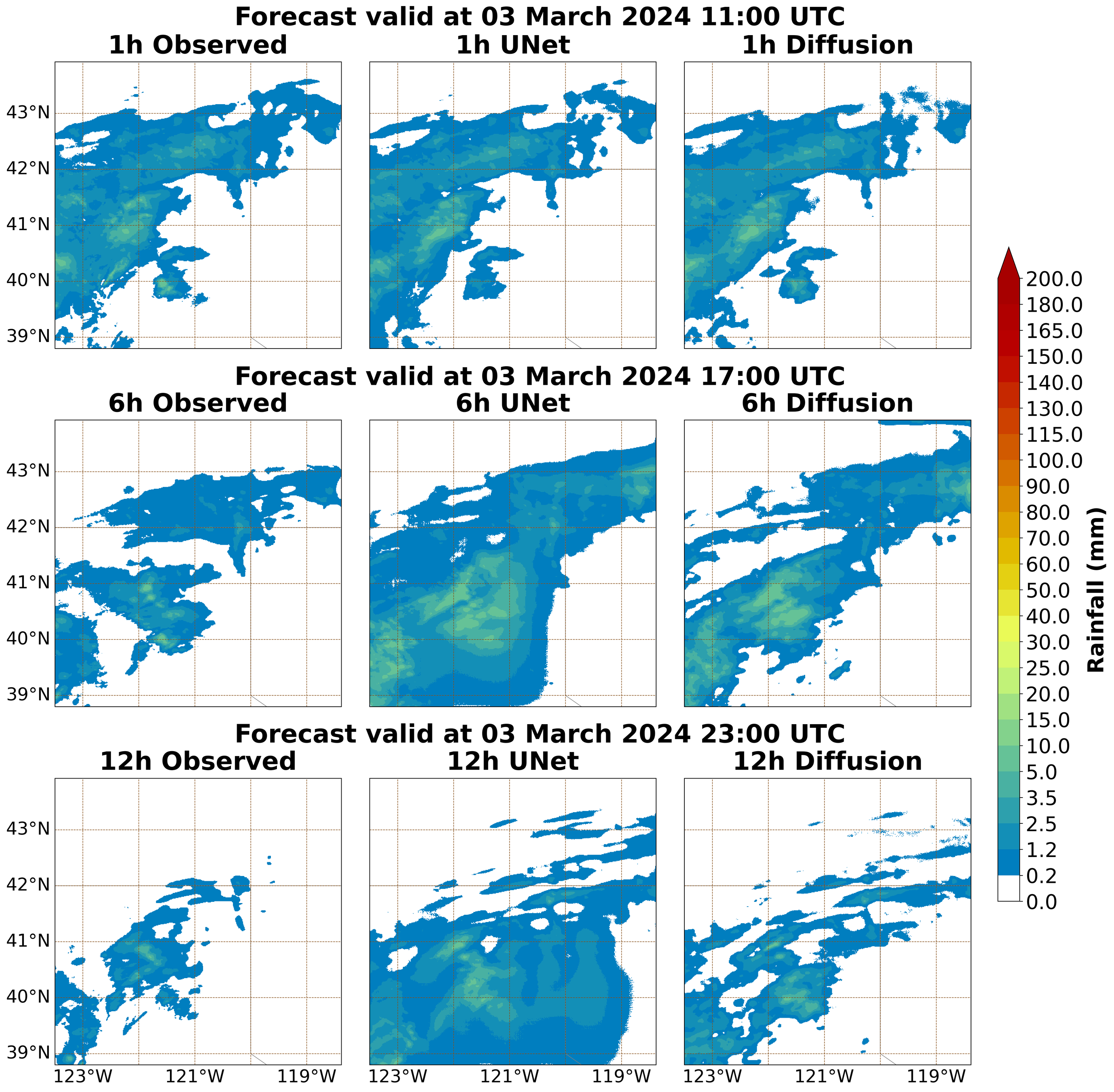}
    \caption{Oregon precipitation forecasts}
    \label{fig:qual_oregon}
\end{subfigure}

\vspace{0.1cm}

\begin{subfigure}{\textwidth}
    \centering
    \includegraphics[width=\textwidth]{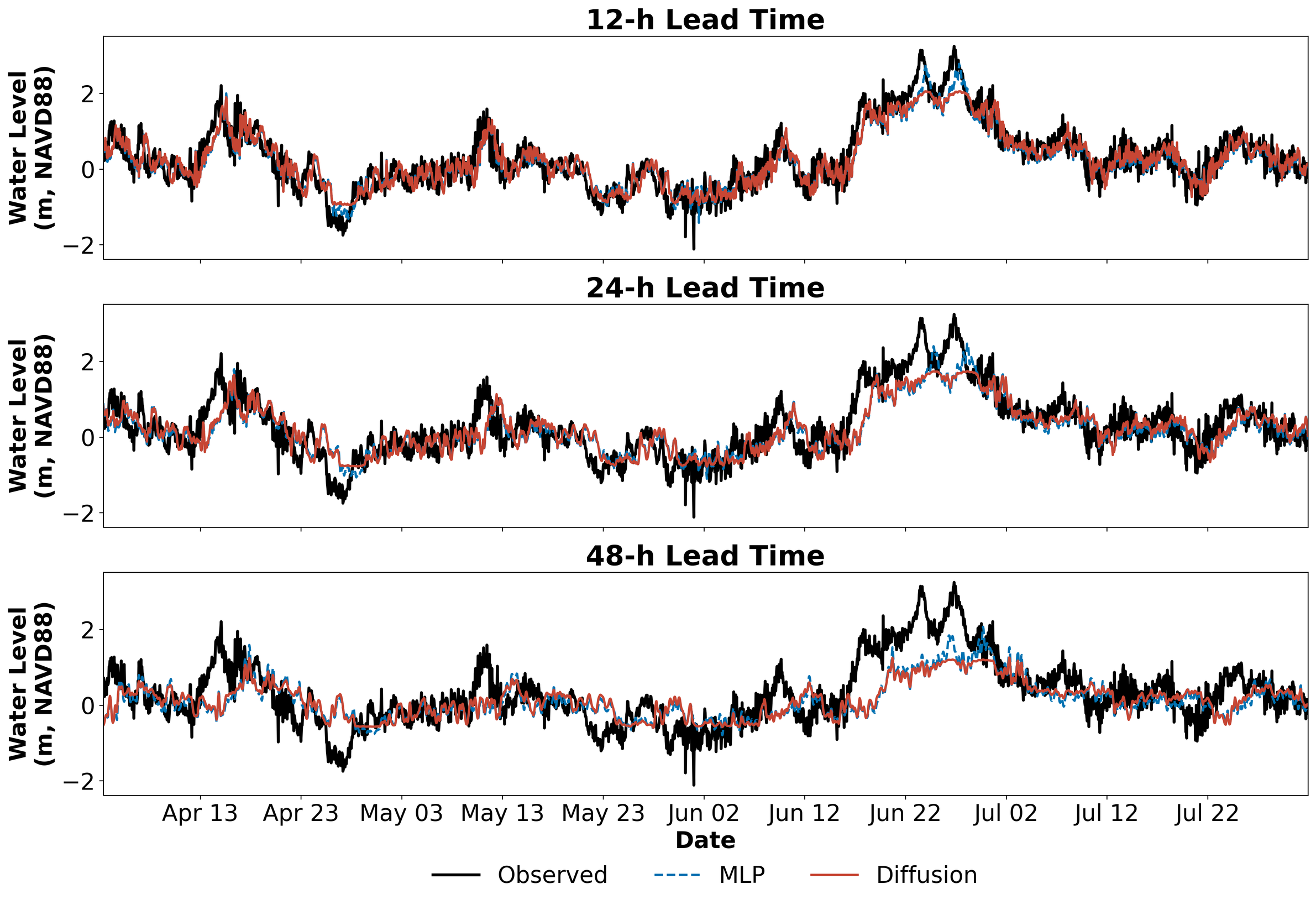}
    \caption{Bob Hall Pier water-level forecasts}
    \label{fig:qual_bhp}
\end{subfigure}

\vspace{0.1cm}

\caption{Representative deterministic and diffusion forecasts. 
(a) Oregon-region precipitation forecasts at 1-, 6-, and 12-hour lead times. 
(b) Bob Hall Pier water-level forecasts at 12-, 24-, and 48-hour lead times. 
The examples illustrate preservation of spatial structure in the raster diffusion forecasts and progressive loss of variability in the tabular forecasts.}
\label{fig:qual_examples}
\end{figure}

These regional differences are important because the tabular models use only current and lagged water-level information. As the rollout progresses, observed inputs are replaced by model predictions until the forecast becomes fully autoregressive and contains no information about future meteorological forcing. Consequently, future water-level changes driven by external forcing that is not inferable from the preceding surge history cannot be recovered by either forecasting approach. The particularly poor Rockport trajectories therefore illustrate an important limit of rollout stabilization. Diffusion can suppress recursive error amplification, but it cannot compensate for missing predictive information. The variation across the three stations further demonstrates that the usefulness of recursive forecasting depends strongly on the predictive information contained in the available surge history and on unresolved external forcing. Figure~\ref{fig:qual_examples}b illustrates how this behavior evolves with lead time for Bob Hall Pier. At the initial 12-hour lead time, both models closely follow the observations, but forecast fidelity progressively decreases as the rollout extends. Both models lose amplitude and event-level fidelity, with diffusion exhibiting stronger contraction toward central values. Corresponding lead-time examples for Port Isabel and Rockport are provided in~\ref{appendix:visual_examples}.

\subsection{Raster Rollout Evaluation}

\begin{figure}[!htbp]
\begin{center}
 \noindent\includegraphics[width=\textwidth]{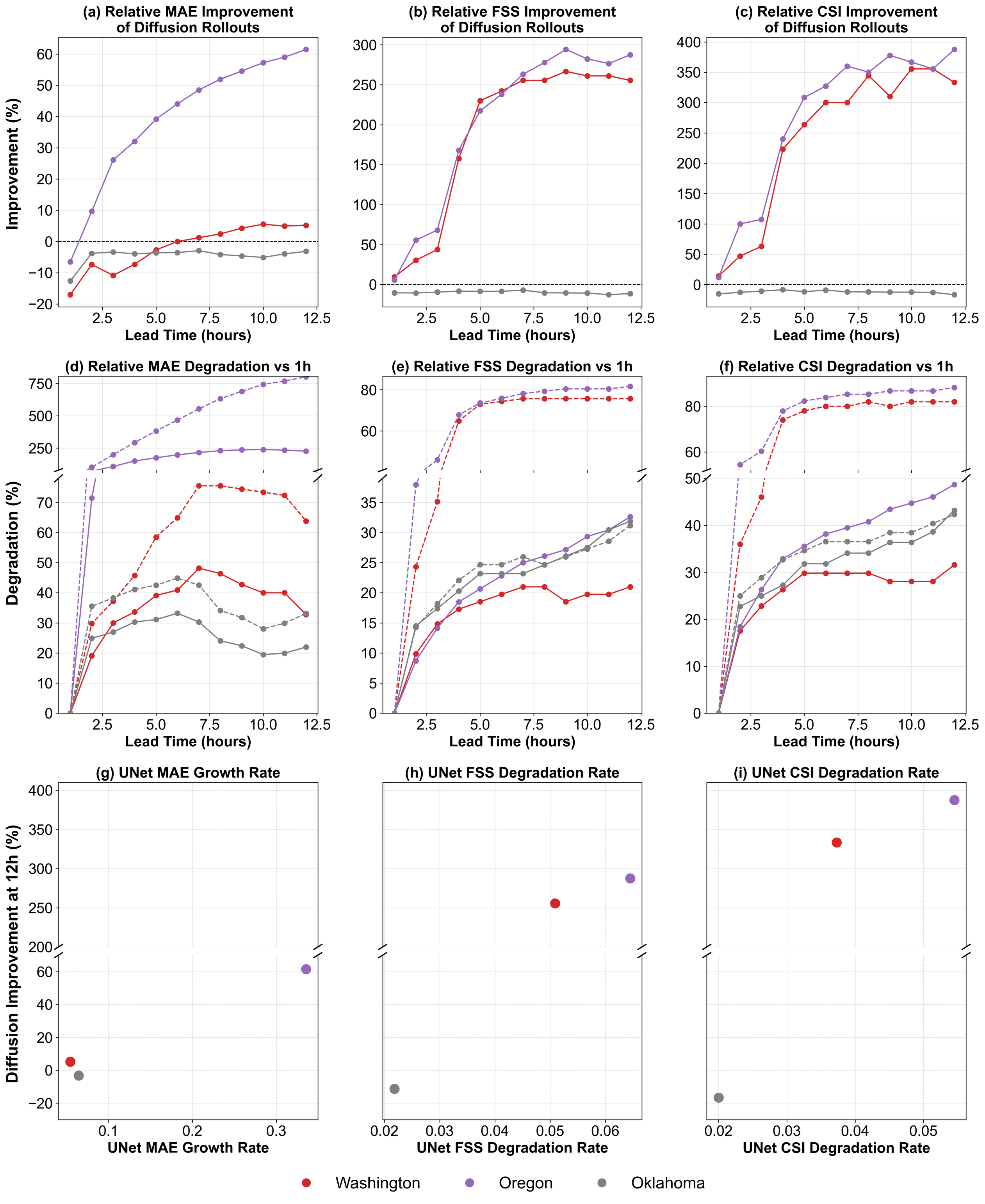}\\
\caption{Raster rollout performance across the Washington, D.C., Oregon, and Oklahoma regions. (a--c) Relative MAE, FSS, and CSI improvement of diffusion rollouts compared with deterministic U-Net rollouts across lead times. Positive values indicate improved diffusion performance. (d--f) Relative degradation in MAE, FSS, and CSI with increasing lead time for diffusion (solid) and deterministic U-Net (dashed) rollouts. (g--i) Relationship between deterministic rollout instability and diffusion improvement at 12-hour lead time. Larger deterministic degradation corresponds to greater relative diffusion stabilization.}
 \label{fig_raster_results}
\end{center}
\end{figure}

Figures~\ref{fig_raster_results}a--c show the relative performance of diffusion rollouts compared with the deterministic U-Net across the three raster regions. Diffusion provides the clearest improvements in the Washington, D.C. and Oregon regions, particularly at longer lead times. In contrast, the Oklahoma region shows little improvement in MAE. Importantly, FSS and CSI exhibit substantially larger diffusion gains than MAE in Washington, D.C. and Oregon, indicating that the principal benefit is preservation of precipitation structure and event-detection skill rather than uniformly improved pointwise magnitude. Figures~\ref{fig_raster_results}d--f further show that diffusion generally slows degradation in MAE, FSS, and CSI as lead time increases. The largest differences occur where deterministic structural skill deteriorates most rapidly. Even in Oklahoma, where absolute MAE improvement is limited, diffusion moderates the rate of structural degradation relative to the deterministic rollout. Figures~\ref{fig_raster_results}g--i similarly show that stronger deterministic degradation is associated with larger relative diffusion stabilization, consistent with the tabular results.

The representative Oregon forecast in Figure~\ref{fig:qual_examples}a illustrates the spatial manifestation of these metric differences. As lead time increases, the deterministic U-Net develops broader and less coherent precipitation features, whereas the diffusion forecast retains greater spatial organization in this example. This qualitative behavior is consistent with the larger FSS and CSI improvements relative to MAE, which provide the quantitative evidence for improved preservation of precipitation structure. Corresponding examples for Washington, D.C. and Oklahoma are provided in~\ref{appendix:visual_examples}.

Together, the experiments distinguish two sources of long-horizon forecast degradation. Recursive error amplification occurs when model-generated states are repeatedly fed back into the forecasting system, and diffusion can substantially moderate this instability. Forecast degradation can also arise when information needed to predict future evolution is absent from the conditioning variables. Diffusion cannot recover information that is unavailable to the model. The water-level experiments eventually become fully dependent on previous model predictions, whereas the precipitation experiments continue to receive advancing HRRR forecasts throughout recursion.

The contrasting behavior across these settings indicates that diffusion-based stabilization is most informative when interpreted together with the predictive information available during rollout. In the HRRR-conditioned precipitation experiments, improved stability is accompanied by better preservation of spatial structure and event-detection skill. In the water-level experiments, and most clearly at Rockport, diffusion can remain numerically bounded even after substantial event-level information has been lost. These results therefore support treating rollout stability and forecast fidelity as complementary but distinct properties of long-horizon environmental forecasts.

\section{Conclusions}
This study evaluates diffusion-based rollouts as a stabilization mechanism for long-horizon recursive environmental forecasting across water-level time series and spatial precipitation fields. Across both forecasting settings, diffusion moderates recursive error growth and limits the degradation associated with repeatedly feeding model predictions back into subsequent forecasts. The magnitude and practical value of this stabilization, however, vary substantially across forecasting regimes.

The water-level experiments demonstrate an important limitation of recursive stabilization. Because these forecasts rely only on current and lagged water-level information, they progressively lose access to independent predictive information as the rollout becomes fully autoregressive. The three tide gauge stations exhibit distinct behaviors under this constraint. Bob Hall Pier remains comparatively stable, Port Isabel shows stronger loss of event amplitude, and Rockport illustrates a regime in which both deterministic and diffusion forecasts lose substantial event-level fidelity. In such cases, diffusion can prevent rapid numerical divergence but cannot recover future evolution driven by external forcing that is absent from the model inputs. Numerical stability therefore does not necessarily imply an informative forecast.

The precipitation experiments provide a complementary forecasting regime in which advancing HRRR forecasts continue to supply information about future atmospheric evolution throughout recursion. In this setting, diffusion stabilization is associated with better preservation of coherent precipitation structure and event-detection skill, particularly where deterministic rollouts degrade rapidly.

Together, these results distinguish two limitations of long-horizon recursive forecasting. These are error amplification during recursion and loss of predictive information. Diffusion directly addresses recursive instability but cannot recover information that is absent from the model inputs. The contrasting behavior across the evaluated forecasting settings indicates that the practical benefit of diffusion depends on both the degree of recursive degradation and the information available to constrain future evolution. More broadly, rollout stability and forecast fidelity should be evaluated separately when assessing long-horizon machine-learning forecasts.

\appendix

\section{Appendix A: Dataset Splits}
\label{appendix:years}

This appendix summarizes the training, validation, and testing years used for each NOAA tide gauge station in the low-dimensional tabular experiments. These datasets were originally introduced by \cite{vicens2024exploring}, where the years were selected based on data completeness and quality.

\begin{table}[H]
\centering
\caption{Training, validation, and testing years for each NOAA tide gauge station.}
\label{tab:A1}
\begin{tabular}{lccc}
\hline
Station & Training Years & Validation Year & Test Year \\
\hline
Bob Hall Pier & 2008--2010 & 2011 & 2012 \\
Port Isabel & 2007, 2009--2010 & 2011 & 2012 \\
Rockport & 2009--2011 & 2012 & 2013 \\
\hline
\end{tabular}
\end{table}

\section{Appendix B: Diffusion Hyperparameters}
\label{appendix:diffusion_hparams}

This appendix summarizes the diffusion hyperparameters used for each tide gauge location in the low-dimensional experiments. The diffusion framework and MLP backbone are consistent across locations, while selected diffusion hyperparameters are specified separately for each station.

Following the EDM formulation of \cite{karras2022elucidating}, $P_{\text{mean}}$ and $P_{\text{std}}$ define the mean and standard deviation of the log-normal distribution used to sample diffusion noise levels during training. The parameter $\sigma_{\text{data}}$ represents the assumed standard deviation of the clean training data used in EDM preconditioning. The parameter $p_{\text{det}}$ is an additional training hyperparameter used in the tabular diffusion configuration.

\begin{table}[H]
\centering
\caption{Diffusion hyperparameters used for each tide gauge location.}
\label{tab:B1}
\begin{tabular}{lcccc}
\hline
Location & $P_{\text{mean}}$ & $P_{\text{std}}$ & $\sigma_{\text{data}}$ & $p_{\text{det}}$ \\
\hline
Bob Hall Pier & -3.86 & 0.5 & 0.2 & 0.2 \\
Port Isabel & -3.86 & 0.5 & 0.2 & 0.2 \\
Rockport & -3.86 & 0.75 & 0.2 & 0.2 \\
\hline
\end{tabular}
\end{table}

\section{Appendix C: Raster Forecasting Architecture}
\label{appendix:architecture}

This appendix summarizes the predictive backbone used in the high-dimensional raster experiments. The dilated attention U-Net follows the architecture introduced by \cite{vicens2025diffusion}. The same predictive backbone is used for the deterministic and diffusion configurations, with the latter embedded within the EDM framework.

\begin{figure}[!htbp]
\centering
\includegraphics[width=\textwidth]{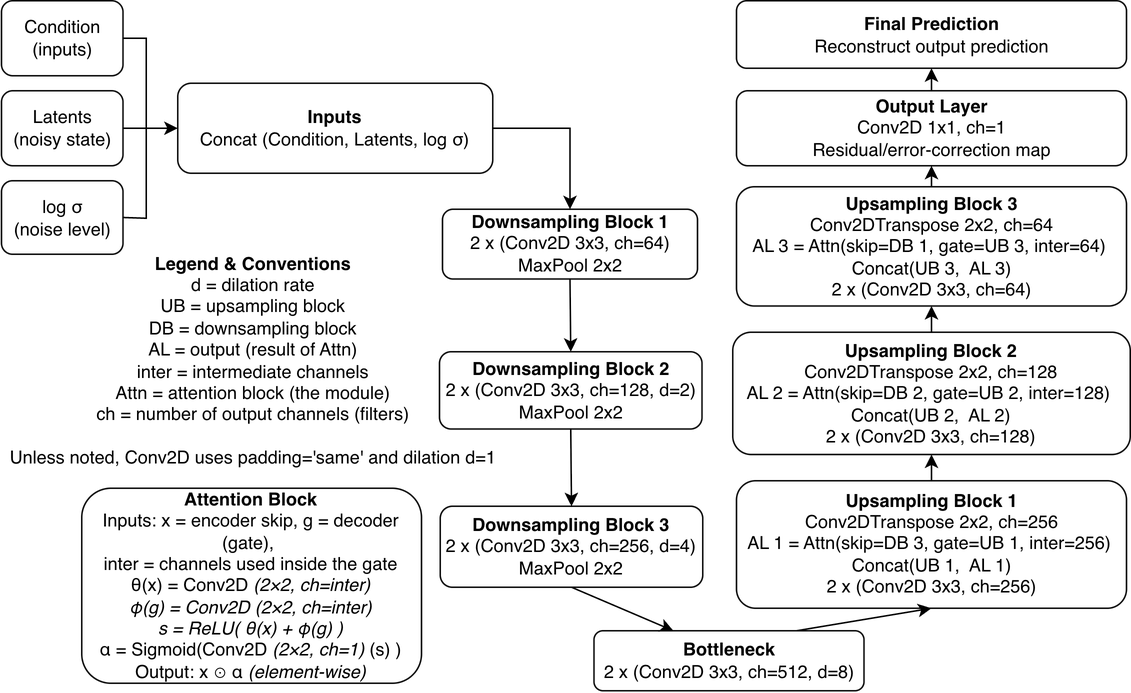}
\caption{Dilated attention U-Net architecture used for raster forecasting, following \cite{vicens2025diffusion}. In the diffusion configuration, the predictive backbone is embedded within the EDM framework \citep{karras2022elucidating}.}
\label{fig:appendix_architecture}
\end{figure}

\section{Appendix D: Additional Visual Examples}
\label{appendix:visual_examples}

This appendix provides additional qualitative examples for the Washington, D.C. and Oklahoma precipitation raster regions and the Port Isabel and Rockport tide gauge stations. Together with the examples in the main text, these forecasts illustrate how diffusion-based stabilization manifests across regions with different degrees of recursive degradation. The raster examples complement the quantitative MAE, FSS, and CSI results by showing differences in precipitation structure as lead time increases, while the water-level examples illustrate that numerical stabilization can occur alongside progressive loss of event-level fidelity.

\begin{figure}[!htbp]
\centering

\begin{subfigure}{\textwidth}
    \centering
    \includegraphics[width=0.7\textwidth]{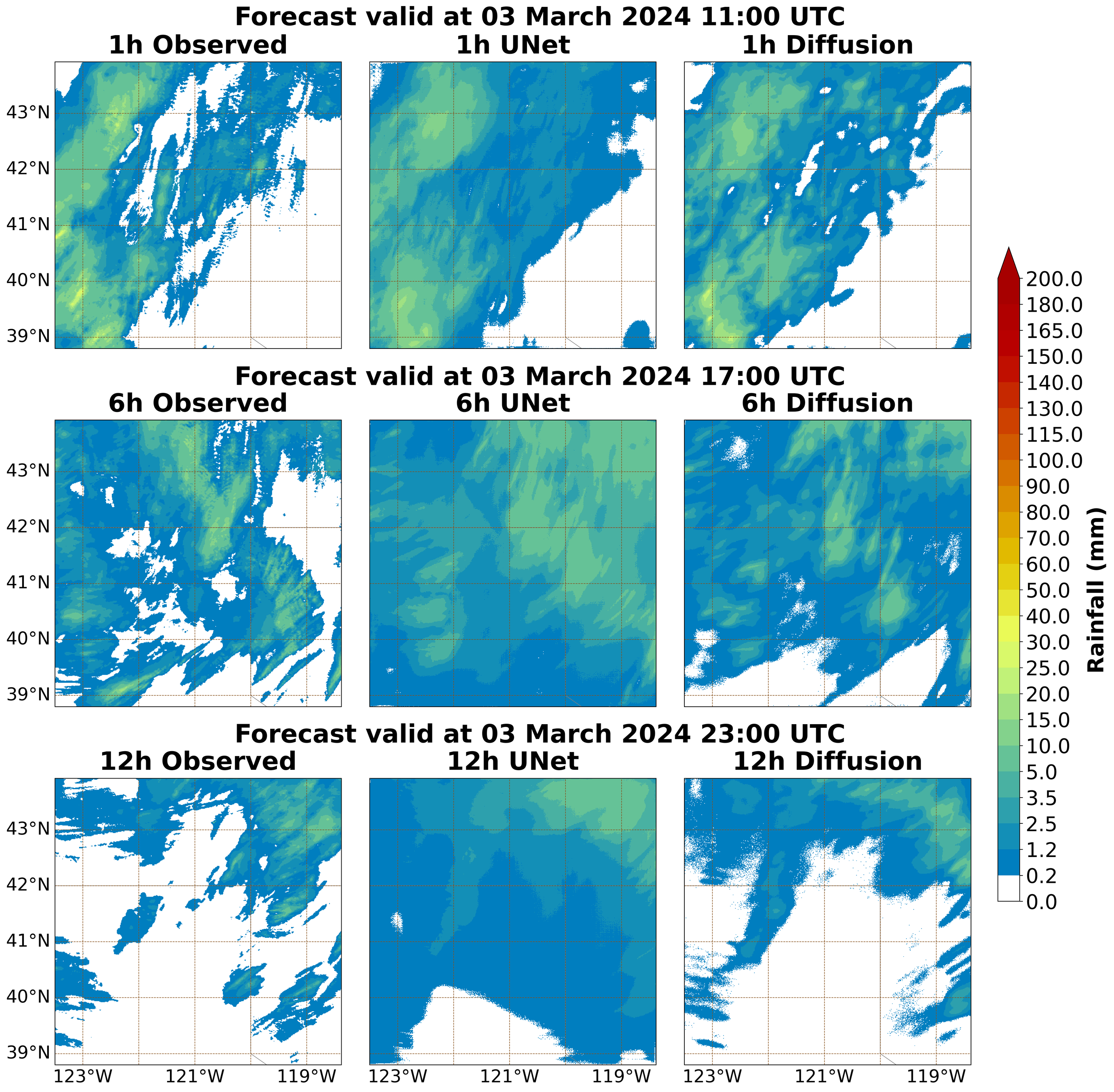}
    \caption{Washington, D.C. precipitation forecasts}
    \label{fig:appendix_washington_precip}
\end{subfigure}

\vspace{0.1cm}

\begin{subfigure}{\textwidth}
    \centering
    \includegraphics[width=\textwidth]{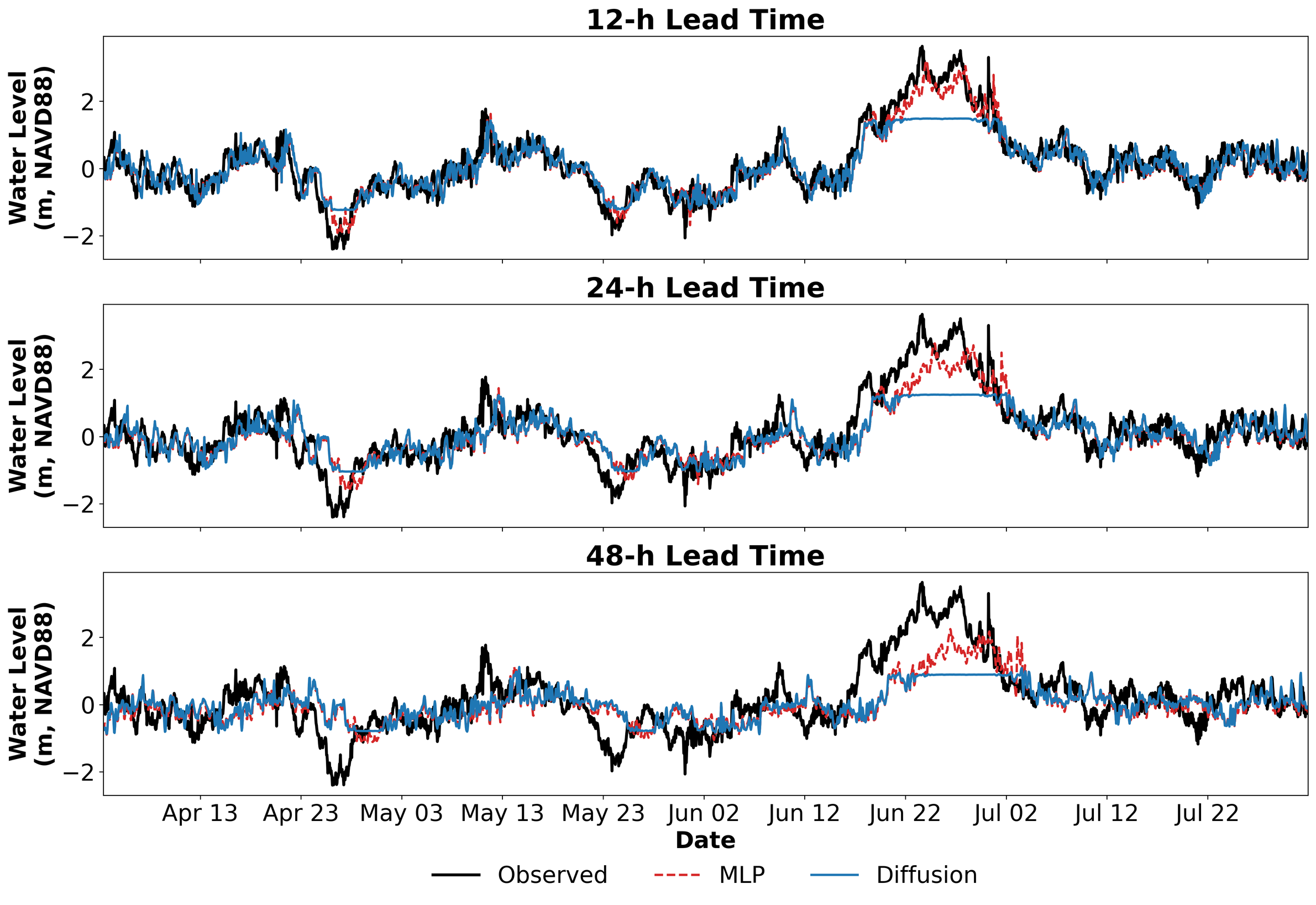}
    \caption{Port Isabel water-level forecasts}
    \label{fig:appendix_port_isabel}
\end{subfigure}

\vspace{0.1cm}

\caption{Additional deterministic and diffusion forecast examples for the Washington, D.C. raster region and Port Isabel tide gauge. In Washington, D.C., diffusion retains greater spatial organization as lead time increases, whereas the deterministic U-Net develops broader and smoother precipitation features. At Port Isabel, both approaches progressively lose event amplitude and variability with increasing lead time, illustrating that reduced recursive degradation does not necessarily preserve event-level fidelity.}
\label{fig:appendix_washington_portisabel}

\end{figure}

\begin{figure}[!htbp]
\centering

\begin{subfigure}{\textwidth}
    \centering
    \includegraphics[width=0.7\textwidth]{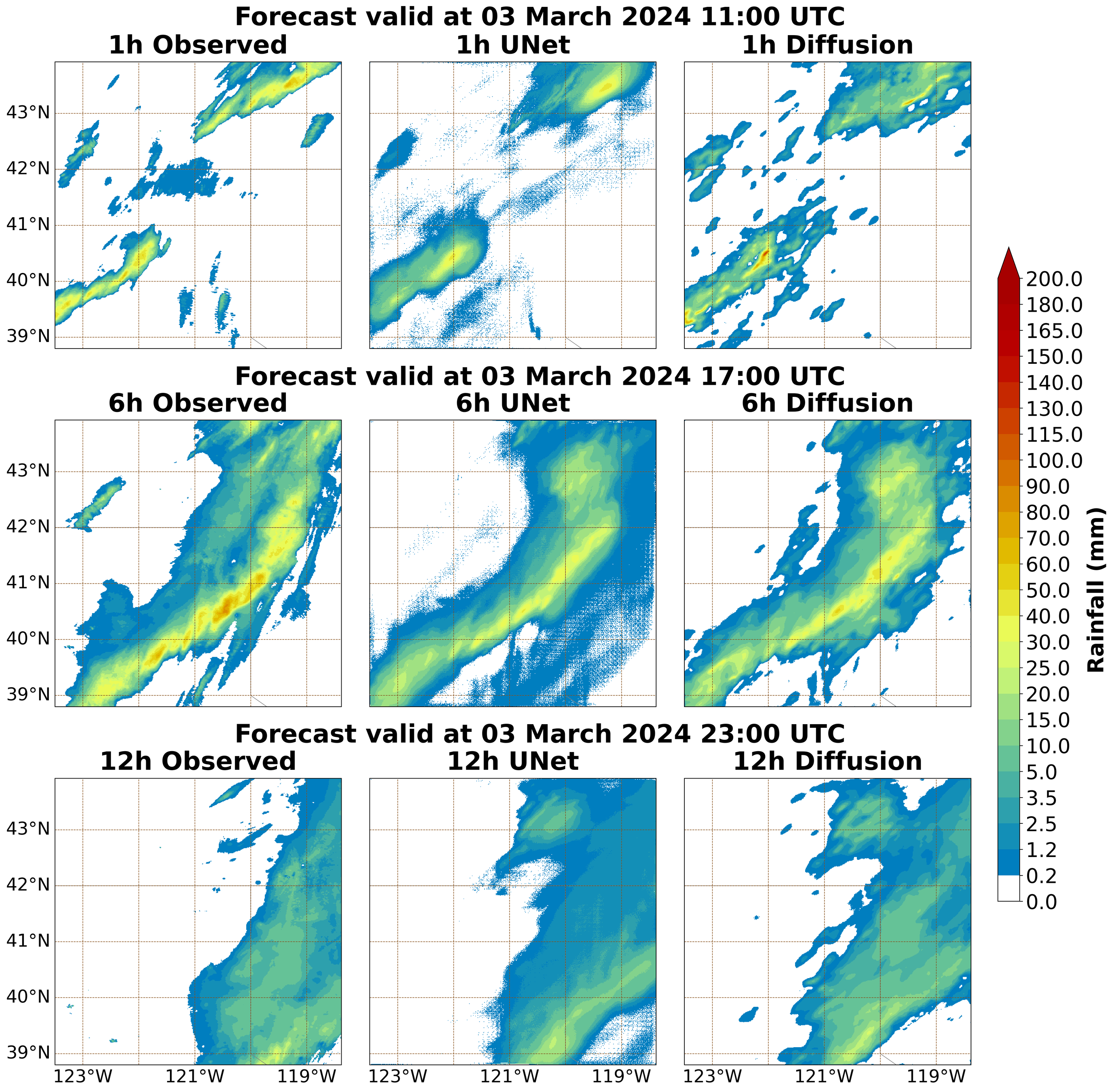}
    \caption{Oklahoma precipitation forecasts}
    \label{fig:appendix_oklahoma_precip}
\end{subfigure}

\vspace{0.1cm}

\begin{subfigure}{\textwidth}
    \centering
    \includegraphics[width=\textwidth]{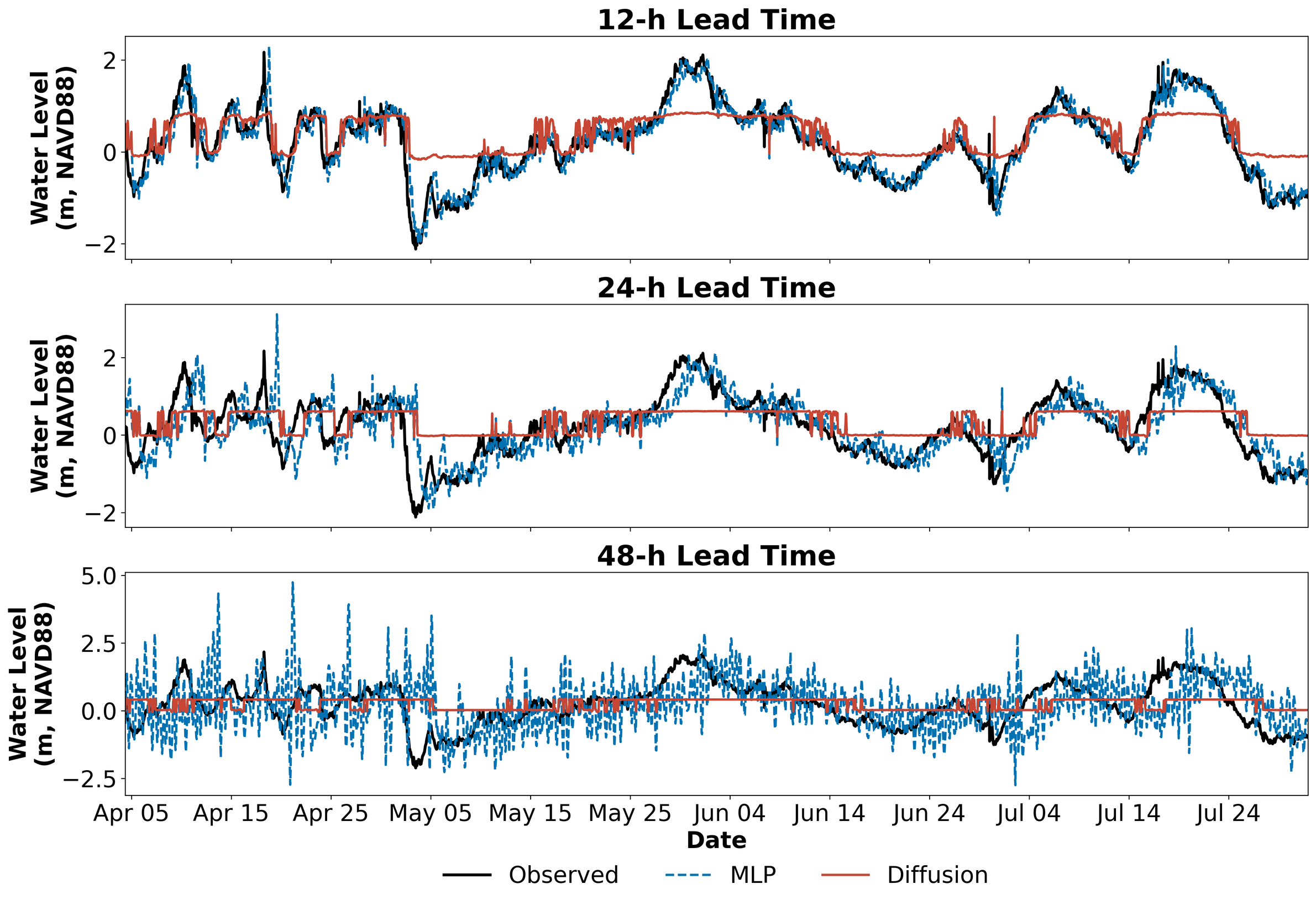}
    \caption{Rockport water-level forecasts}
    \label{fig:appendix_rockport}
\end{subfigure}

\vspace{0.1cm}

\caption{Additional deterministic and diffusion forecast examples for the Oklahoma raster region and Rockport tide gauge. In Oklahoma, differences between the deterministic and diffusion forecasts are more modest than in the other raster regions, consistent with the smaller quantitative diffusion gains. Rockport provides the contrasting water-level case, where diffusion remains numerically bounded while substantial event-level fidelity is lost, demonstrating that rollout stabilization alone does not guarantee an informative forecast.}
\label{fig:appendix_oklahoma_rockport}

\end{figure}

\section*{Open Research Section}

The NOAA tide gauge water-level datasets used in the tabular experiments are publicly available through the repository associated with \cite{vicens2024exploring} at \url{https://github.com/conrad-blucher-institute/waterLevelJournal}. NOAA Multi-Radar Multi-Sensor (MRMS) precipitation observations and High-Resolution Rapid Refresh (HRRR) forecasts used in the raster experiments are publicly available from NOAA data services. The preprocessing procedures and raster dataset construction follow \cite{vicens2025diffusion}.

Code required to reproduce the deterministic and diffusion model configurations,
recursive rollout procedures, and evaluation presented in this study is publicly
available at \url{https://github.com/ai2es/diffusion_rollout_stability}.

\section*{Conflict of Interest}
The authors declare there are no conflicts of interest for this manuscript.

\acknowledgments
This material is based upon work supported by the National Science Foundation under Grant No. RISE-2019758 within the NSF AI Institute for Research on Trustworthy AI in Weather, Climate, and Coastal Oceanography (AI2ES), and under Grant No. IIS2324008. Any opinions, findings, and conclusions or recommendations expressed in this material are those of the author(s) and do not necessarily reflect the views of the National Science Foundation. 

The computing for this project was performed at the OU Supercomputing Center for Education \& Research (OSCER) at the University of Oklahoma (OU).

%
%

\bibliography{agusample}

%
%
%
%
%

\end{document}